\documentclass[10pt,twocolumn,letterpaper]{article}

\usepackage{cvpr}              
\usepackage[accsupp]{axessibility}
\usepackage{graphicx}
\usepackage{amsmath}
\usepackage{amssymb}
\usepackage{booktabs}
\usepackage{makecell}
\usepackage{multicol}
\usepackage{multirow}
\usepackage{float}
\usepackage{bm}

\usepackage[pagebackref,breaklinks,colorlinks]{hyperref}

\usepackage[capitalize]{cleveref}
\crefname{section}{Sec.}{Secs.}
\Crefname{section}{Section}{Sections}
\Crefname{table}{Table}{Tables}
\crefname{table}{Tab.}{Tabs.}

\def\cvprPaperID{8388} 
\def\confName{CVPR}
\def\confYear{2022}
\usepackage{color}

\begin{document}

\title{Personalized Image Aesthetics Assessment with Rich Attributes}

\author {Yuzhe Yang$^1$\footnotemark[1] ,
Liwu Xu$^1$\footnotemark[1] ,
Leida Li$^2$, 
Nan Qie$^1$, 
Yaqian Li$^1$,
Peng Zhang$^1$,
Yandong Guo$^1$\footnotemark[2]\\
$^1$OPPO Research Institute \quad
$^2$Xidian University\\
{\tt\small ippllewis@gmail.com, \{xuliwu, qienan, liyaqian, zhangpeng6\}@oppo.com}\\
{\tt\small ldli@xidian.edu.cn, yandong.guo@live.com}
}
\maketitle

\renewcommand{\thefootnote}{\fnsymbol{footnote}}
\footnotetext[1]{Equal contribution}
\footnotetext[2]{Corresponding author}

\begin{abstract}
Personalized image aesthetics assessment (PIAA) is challenging due to its highly subjective nature. People's aesthetic tastes depend on diversified factors, including image characteristics and subject characters. The existing PIAA databases are limited in terms of annotation diversity, especially the subject aspect, which can no longer meet the increasing demands of PIAA research. To solve the dilemma, we conduct so far, the most comprehensive subjective study of personalized image aesthetics and introduce a new Personalized image Aesthetics database with Rich Attributes (PARA), which consists of 31,220 images with annotations by 438 subjects. PARA features wealthy annotations, including 9 image-oriented objective attributes and 4 human-oriented subjective attributes. In addition, desensitized subject information, such as personality traits, is also provided to support study of PIAA and user portraits. A comprehensive analysis of the annotation data is provided and statistic study indicates that the aesthetic preferences can be mirrored by proposed subjective attributes. We also propose a conditional PIAA model by utilizing subject information as conditional prior. Experimental results indicate that the conditional PIAA model can outperform the control group, which is also the first attempt to demonstrate how image aesthetics and subject characters interact to produce the intricate personalized tastes on image aesthetics. We believe the database and the associated analysis would be useful for conducting next-generation PIAA study. The project page of PARA can be found at: \url{https://cv-datasets.institutecv.com/#/data-sets} .
\end{abstract}

\section{Introduction}
\label{sec:intro}

\begin{figure}[h]
    \centering
    \includegraphics[width=0.46\textwidth]{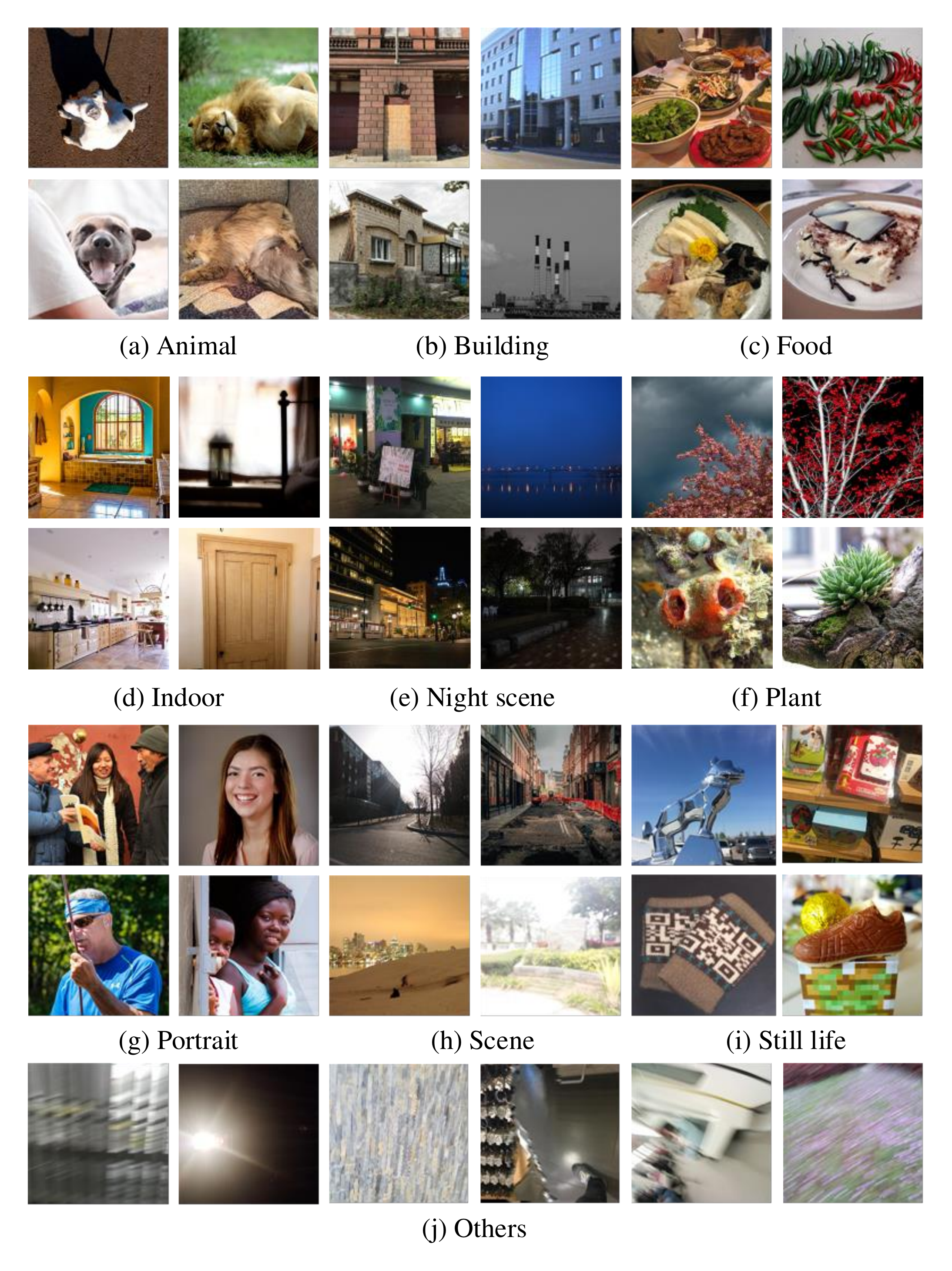}
    \caption{Sample images in PARA.}
    \label{fig1}
\end{figure}

Image aesthetics assessment (IAA) aims at evaluating photo aesthetics computationally. Due to the highly differentiated aesthetic preference, image aesthetics assessment can be divided into two categories: generic and personalized image aesthetics assessment (a.k.a GIAA and PIAA) \cite{Ren_2017_ICCV}. For GIAA, an image is annotated by different voters and mean opinion score (MOS) is used as the aesthetics “ground truth”. However, GIAA merely reflects an ``average opinion", which neglects the highly subjective nature of aesthetic tastes. To mitigate this issue, PIAA was proposed to capture unique aesthetic preferences \cite{Ren_2017_ICCV}.
In the past decade, PIAA has achieved encouraging advances. Initially, Ren et al. \cite{Ren_2017_ICCV} proposed the first PIAA database named FLICKR-AES and they addressed this problem by leveraging GIAA knowledge on user-related data, so that model can capture aesthetic ``offset". Later, research work attempted to learn PIAA from various perspectives, such as multi-modal collaborative learning \cite{wang2018collaborative}, meta-learning \cite{zhu2020personalized}, multi-task learning \cite{li2020personality}, deep reinforcement learning \cite{lv2021user} etc. High quality databases are essential for building data-driven PIAA models. However, current databases, such as FLICKR-AES \cite{Ren_2017_ICCV} and AADB \cite{kong2016aesthetics}, are limited in annotation diversity. For comparison, we summarize the annotation information of three related databases in Table \ref{tab:comparison} and it is easy to observe that most databases are limited in annotation diversity.

\begin{table}[h]
  \centering
  \resizebox{0.47\textwidth}{!}{
  \begin{tabular}{l|cc|ccccc}
    \toprule
        \toprule
    Database & \makecell[c]{Subjective \\label} & \makecell[c]{Objective\\ label} & \makecell[c]{Annotation \\count}& \makecell[c]{Avg. annotation \\times per image}  & \makecell[c]{Num. of \\images} &  \makecell[c]{Num. of annotation\\ dimension} &\makecell[c]{Num. of \\ subject}\\
    \midrule
    AADB \cite{kong2016aesthetics} &  & \checkmark &    600 k  & 5  & 10,000            & 12    & 190\\
    REAL CUR \cite{Ren_2017_ICCV}      &  & \checkmark  &   2.87 k & 1 & 2,870              & 1    & 14\\
    FLICKR-AES \cite{Ren_2017_ICCV}    &  & \checkmark   & 200 k  & 5  & \textbf{40,000}   & 1     & 210\\
    \midrule
    \textbf{Ours (PARA)}      & \checkmark  & \checkmark & \textbf{$ \sim $ 9723 k} & \textbf{25.87} & 31,220            & \textbf{13}  &\textbf{438}\\
    \bottomrule
    \bottomrule
  \end{tabular}}
  \caption{Comparison among PIAA databases. Note that the ``annotation count" is calculated by multiply Num. of images, average annotation times per image and Num. of annotation dimension. Since scene label is assigned to each image before annotation begins, we add the number of scene label separately.}
  \label{tab:comparison}
\end{table}

\begin{table}[htbp]
  \centering
  \resizebox{0.47\textwidth}{!}{
  \begin{tabular}{llll}
     \toprule
    \toprule
    \multicolumn{2}{c}{\textbf{Objective label}} & \multicolumn{2}{c}{\textbf{\# Subjective label}} \\
    \midrule
    Session ID              &        session1               &   \textbf{User ID}        &           A3c6418  \\
    Image name              &        iaa\_pub1\_.jpg        &   Age                     &           30  \\
    \textbf{User ID}        &        A3c6418           &   Gender                  &           male  \\
    Aesthetics        &        3.0                    &   Education experience         &           University\\
    Quality           &        3.1                    &   Artistic experience     &           proficient  \\
    Composition       &        3                      &   Photographic experience &           proficient  \\
    Color             &        4                      &   E       &        5  \\
    Depth of Field               &        3                      &   A       &        9  \\
    Content           &        3                      &   N       &        4  \\
    Light             &        4                      &   O       &        7  \\
    object emphasis        &        0 (False)                      &   C       &        9  \\
    \cmidrule{3-4}
    Scene categories        &       animal                  & Emotion &         Neutral  \\ 
    
                            &                               & Difficulty of judgement &         -1 (Easy) \\  
                            &                               & Content preference      &         3 (Neutral)   \\
                            &                               & Willingness to share    &         3 (Neutral)    \\
\bottomrule
\bottomrule
  \end{tabular}}
  \caption{Annotation information of single image. The annotations are divided into two groups, including objective and subjective information. The subjective and objective annotation can be associated by user ID.}
  \label{tab:annotation}
\end{table}
To mitigate this issue, we notice that attributes usually provide a richer description to explicitly characterize differentiation \cite{kong2016aesthetics}. Therefore, beyond aesthetics score, we provide quantitative personalized aesthetics attributes annotations to facilitate more accurate aesthetic preference modeling. Here, considering the highly subjective nature of PIAA task, different from the existing databases FLICKR-AES \cite{Ren_2017_ICCV} and AADB \cite{kong2016aesthetics}, we design the label system of PARA from two perspectives, which are human-oriented and image-oriented annotations. Specifically, apart from image aesthetics attributes, we also collect subjective annotations, including 1) content preference, 2) difficulty of judgment, 3) emotion, 4) willingness to share. We believe the aforementioned dimensions can bring further research opportunities in understanding correlation between PIAA and psychological feelings. In addition, we also provide desensitized subject information (user ID, age, gender, education, personality trait, photographic experience, art experience) for more in-depth analysis in the future. Annotation dimensions of single image are demonstrated in Table \ref{tab:annotation}. \par

In this paper, we build so far, the richest annotated personalized image aesthetic assessment database named ``PARA". In addition, we also conduct an in-depth analysis of annotation information and propose a benchmark for this database. Contributions of this work can be summarized as follows:
\begin{itemize}
    \item We conduct so far, the most comprehensive subjective study of personalized image aesthetics, and build a PIAA database with rich annotations. Specifically, we collected 31,220 images and each image is annotated by 25 subjects in average and 438 subjects in total. Each image is annotated with 4 human-oriented subjective attributes and 9 image-oriented objective attributes. To support in-depth analysis, we also provide desensitized subject information.
    \item We provide an in-depth analysis to discover characteristics of annotations dimensions. Statistical results indicate that the personalized aesthetic preference can be mirrored by the proposed human-oriented subjective attributes, including personality traits, difficulty of judgement and image emotion, which in turn enlightens novel research angles, such as modeling personalized aesthetic by utilizing subject information.
    \item We conduct a benchmark study based on the proposed PARA database. The benchmark contains two models, including unconditional and conditional PIAA. By utilizing subject information as a condition when modeling aesthetic preference, we prove that training with human-oriented annotations can further improve the PIAA model performance.

\end{itemize}
\begin{figure*}[ht]
    \centering
    \includegraphics[width=0.91\textwidth]{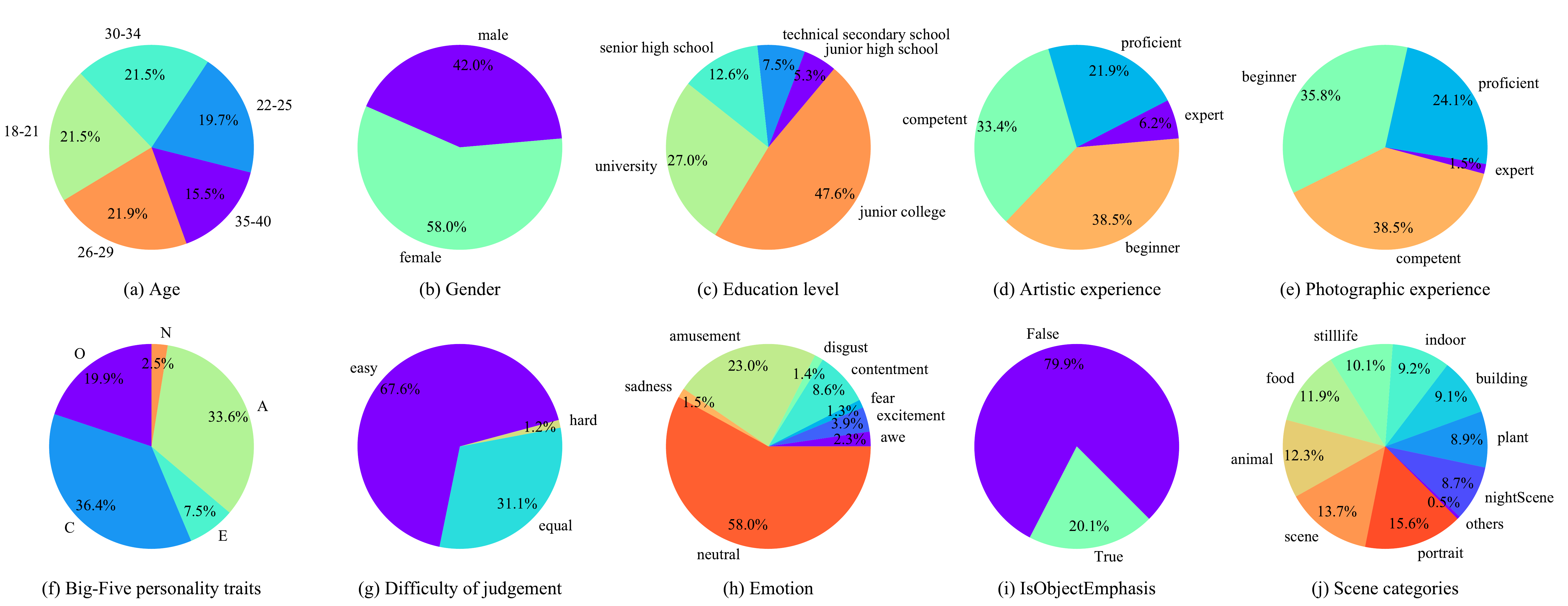}
    \caption{Statistical pie charts of user portraits and attributes in PARA: (a) age, (b) gender, (c) education experience, (d) artistic experience, (e) photographic experience, (f) Big-Five personality traits, (g) difficulty of judgement, (h) emotion distribution, (i) object emphasis, (j) scene categories.}
    \label{fig2}
\end{figure*}

\section{Related Works}
\label{sec:related}
\subsection{Databases for PIAA}
For most data-driven machine learning systems, data with rich annotation plays a critical role. In PIAA research, three databases are frequently used, including FLICKR-AES \cite{Ren_2017_ICCV}, REAL-CUR \cite{Ren_2017_ICCV} and AADB \cite{kong2016aesthetics}.
FLICKR-AES \cite{Ren_2017_ICCV} is actually the first database specially designed for PIAA research. FLICKR-AES contains 40,000 images with a creative commons license from Flickr \footnote{https://www.flickr.com}. It is annotated by 210 AMT \footnote{Amazon Mechanical Turk, https://www.mturk.com/} workers and the aesthetics score ranges from 1 to 5. Higher score indicates better visual aesthetics perception. However, rating scores in FLICKR-AES is provided by AMT workers, instead of the owner of image. To test PIAA algorithms in real-scene, REAL-CUR \cite{Ren_2017_ICCV} was proposed and it is a small-scale database consists of 14 real personal albums. Each album contains different number of photos, ranging from 197 to 222, while the average is 205. In PIAA research, this database is usually served as a test set for algorithm verification only \cite{Ren_2017_ICCV}. In addition to FLICKR-AES and REAL-CUR, another database that usually used in PIAA research is AADB \cite{kong2016aesthetics}. It is initially designed to jointly learn image aesthetics and related attributes. Since subject ID is also provided, AADB can also be utilized to learn personalized aesthetic preferences. AADB contains 10,000 images rated by 190 workers in total and 5 workers in average. AADB provides 11 aesthetics attributes annotation (interesting content, object emphasis, good lighting, color harmony, vivid color, shallow depth of field, motion blur, rule of thirds, balancing element, repetition, and symmetry) and 1 aesthetics score that ranges from 1 to 5, indicating the overall aesthetics judgement.\par
\subsection{PIAA Models}
Different types of computational PIAA models have been derived by various deep learning techniques in the past decade. Ren et al. \cite{Ren_2017_ICCV} proposed the first PIAA databases and they addressed the PIAA task by leveraging GIAA prior knowledge to personalized data to capture individual aesthetic preferences. Specifically, they first trained a GIAA model to provide fundamental task prior. Then, they finetune the GIAA knowledge with attributes and content features to learn a personalized aesthetic ``offset" through residual learning. Wang et al. \cite{wang2018collaborative} pointed out that the current PIAA models contain insufficient user-specific information. Therefore, they enrich the current PIAA database by attaching textual reviews and conducting a user/image relation embedding for collaborative learning. Besides, they introduce an attentive mechanism to dig out image semantic and Region-of-Interest (ROI) via fusing multi-modal annotation information. Zhu et al. \cite{li2020personality} proves that through multitask learning and cross-data training with personality information, performance of both GIAA and PIAA can outperform the other IAA algorithms. More recently, deep meta-learning \cite{vinyals2016matching} has been proved its effectiveness in capturing aesthetic preferences \cite{wang2019meta, zhu2020personalized}. In these works, each user's annotation is regarded as a meta-task. Through the unique episodic training mechanism, the trained model can quickly adapt into new subject data. It is noteworthy that most algorithms mentioned above reflects the necessity and effectiveness of introducing extra information in the learning procedure, which in turn, indicates that extra data are required to design better PIAA models. These promote us to conduct this work and bring the next-generation PIAA.

\section{PARA Database}
The construction of PARA database contains four stages, including data collection, label system design, subject selection and subjective experiments.
\subsection{Data Collection}
We collect images from CC search \footnote{\url{https://search.creativecommons.org/}} and filter images with ``creative commons" license and ``Flickr source" conditions. Then, we use a well-trained scene classification model to automatically predict scene labels on each image. Next, we double-checked the labels and revise the scene annotation manually to maintain annotation quality. Then, we sampled around 28,000 images based on scene label to maintain content diversity. We then add around 3,000 images with clear aesthetics ground truth from a website named Unsplash \footnote{Unsplash, https://unsplash.com/} and image quality assessment databases, including SPAQ \cite{fang2020cvpr} and KonIQ-10K \cite{koniq10k}, to balance aesthetics score distribution.
\subsection{Label system design}
When designing the labeling system of PARA, we refer to both GIAA and PIAA databases jointly \cite{murray2012ava, kong2016aesthetics, Ren_2017_ICCV}. The dimensions of PARA label system are shown in Table \ref{tab:annotation}. Each image is annotated with 13 labels together with subject information. Each dimension is explained below.
\begin{itemize}
    \item Image-oriented attributes scores (composition, light, color, depth of field, object emphasis, content), are mostly discretely annotated from 1 to 5. Specially, the ``object emphasis" is a binary label, which indicates whether there exists a salient object in this image.
    \item Emotion, (including amusement, excitement, contentment, awe, disgust, sadness, fear, neutral), refers to the image emotion \cite{zhao2018affective}. Subjects are allowed to select only one dominant emotion for each image.
    \item Difficulty of judgement, is a discrete label in [-1, 0, +1] and it describes the difficulty pf making judgement on photo aesthetics. ``+1" means difficult, ``0" means normal and ``-1" refers to easy.
    \item Content preference, is a discrete annotation in [1, 5] and it represents the extent of semantic preference. For clear statements and reduce question bias, instead of using the exact expression ``content preference", we choose to use ``I like the content of this photo". The meaning from 1 to 5 refers to ``strongly disagree", ``disagree", ``Neutral", ``agree" and ``strongly agree".
    \item Willingness to share, is a discrete label for social computing and image intent estimation. The original question is ``The willingness of sharing this photo to social media". The meaning from 1 to 5 still refers to ``strongly disagree", ``disagree", ``Neutral", ``agree" and ``strongly agree".
    \item User ID, is designed as a unique common key for associating individual subject information and their annotation records together. Notice that to maintain annotation richness and diversity, there are two sources of subjects participated in this annotation task. For convenience, we distinguish annotation from two sources via alphabet A and B at the beginning of each ID.
    \item Aesthetics score, is a discrete class label ranging from 1 to 5 and it mirrors the comprehensive judgement. To cope with ambiguity, we add a middle choice between each integer scale. A higher score indicates better visual aesthetics perception.
    \item Quality score, represents overall judgement of image quality and it ranges from 1 to 5. Higher score represents better perceptual quality. Worth to mention that in PARA, photo with low perceptual quality contains multiple degradation, including motion blur, JPEG compression and etc.
    \item Scene category, represents the content of this image. We carefully select 9 frequently appeared scenes (including portrait, animal, plant, scene, building, still life, night scene, food and indoor) and 1 ``others" class specially refers to photos without obvious meaning. Note that this label is pre-annotated before the subject experiment begin for keeping content diversity.
\end{itemize}

Beyond the aforementioned annotation information, we also collect desensitized subject information for providing more in-depth research opportunities. Related information includes age, gender, education experience (junior high school, senior high school, Technical secondary school, junior college, and university), personality traits (Here we use the Big-Five personality traits, including Openness (O), Conscientiousness (C), Extroversion (E), Agreeableness (A) and Neuroticism (N)), artistic experience and photographic experience. Here, to help users quickly confirm his / her personality, we use the questionnaire of BFI-10 \cite{RAMMSTEDT2007203} and each subject is asked to finish the questionnaire. We then calculate the score of each personality traits and add it to the annotation data.

\subsection{Subject Selection}
As for the principle of hiring subjects, to maintain quality and diversity of annotation, we hire and select subjects mainly based on four perspectives, health state, working experience, personality traits, and subject portrait. First, we make sure that every subject is in a good health state and no forced work is permitted. To maintain the quality of annotation, all subjects are required to work around data annotation over half-years (full-time or part-time) and their works are qualified in other annotation tasks. Second, we make sure that the subjects' portraits are diverse enough, in terms of age, gender, education, photographic experience and personality traits. According to the previous research conclusions proposed by Zhu et al.  \cite{li2020personality}, we believe that personality traits include important information to capture aesthetic preference. Hence, we specially care the distribution of personality traits. Finally, all subjects should pass the Ishihara color blindness test \cite{Hardy:45}. The user portrait distribution of subjects is shown in Figure \ref{fig2} from (a) to (f). All subjects are aware of the usage of PARA. Subjects disagree with the data usage can require us to delete their annotations and quit the experiment freely.
\subsection{Subjective Experiments}
We conduct subject experiments to collect voting results by following the generic psychological experiment protocol \cite{1709988}. First, we split the whole database into 446 sessions for conducting subject experiments. Each session contains 70 unlabeled images, 5 standard images (a small group of pre-annotated data) and 5 repeated images (images require to be annotated two times to test annotation consistency), to control the annotation quality and consistency. Then, we develop a web-based annotation tool and assign personal accounts and passwords to subjects. Finally, relying on a carefully designed annotation quality control strategy \footnote{Details are provided in the supplementary material.}, all qualified annotations are automatically stored without manual acceptance. Note that not all labels mentioned above are annotated in the subject experiment, such as scene label. It is obtained before the subject experiment begin to keep rich content diversity and balance the scene distribution.

\begin{figure}[htb]
    \centering
    \includegraphics[width=0.4\textwidth]{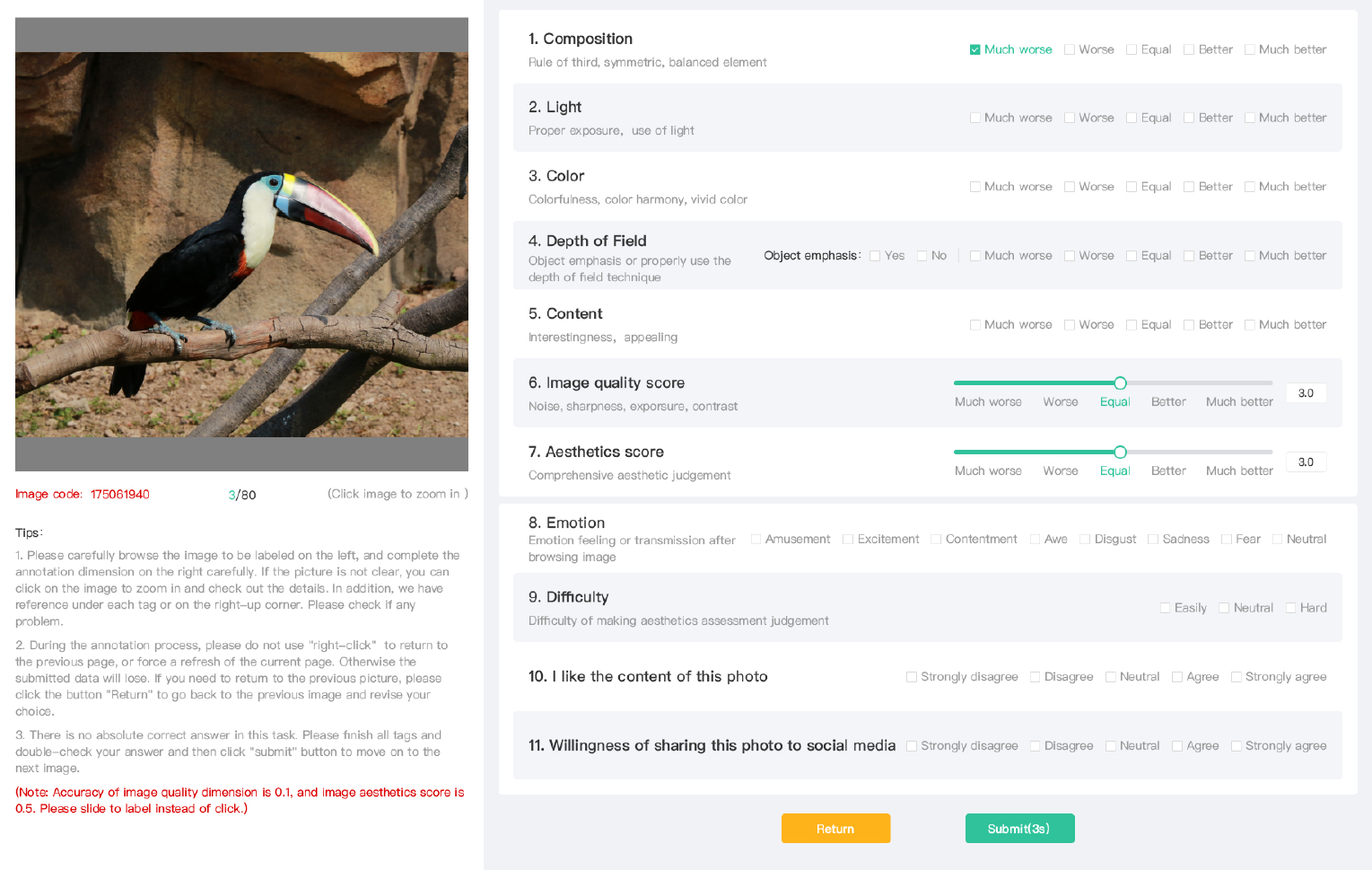}
    \caption{Graphical interface of annotation platform.}
    \label{fig3}
\end{figure}

Figure \ref {fig3} is the graphical interface of annotation platform and subjects are required to fill in all blanks before submitting. To maintain understanding of each dimension, we give unguided explanations below. Subjects can revise annotations by clicking ``return" button.

\section{Data Analysis}
In this section, we first give a summary of the proposed PARA database. Then, we study the aesthetics attributes to discover characteristics of subjective and objective annotations from statistical and correlation perspectives.

\subsection{Data Summary}
The pre-processed PARA contains 31,220 images, with votes from 438 subjects in total. For intuitive observation, we give a group of pie charts in Figure \ref{fig2} to demonstrate the proportion of each dimension. In the first row, pie charts from (a) to (e) demonstrate subject portraits of (a) age, (b) gender, (c) education experience, (d) artistic experience and (e) photographic experience. The second row includes f) Big-Five personality traits, g) difficulty of judgement, (h) emotion, (i) object emphasis (refers to whether there exists a salient object in image), (j) scene categories. Note that in scene categories, we pre-define 9 generic scenes and assign nearly $10\%$ amount of data for each scene to pre-balance the feature diversity of PARA. As for unclear and meaningless images, we assign those with the label ``others" and label them with the rest together.

\begin{figure}[htb]
    \centering
    \includegraphics[width=0.45\textwidth]{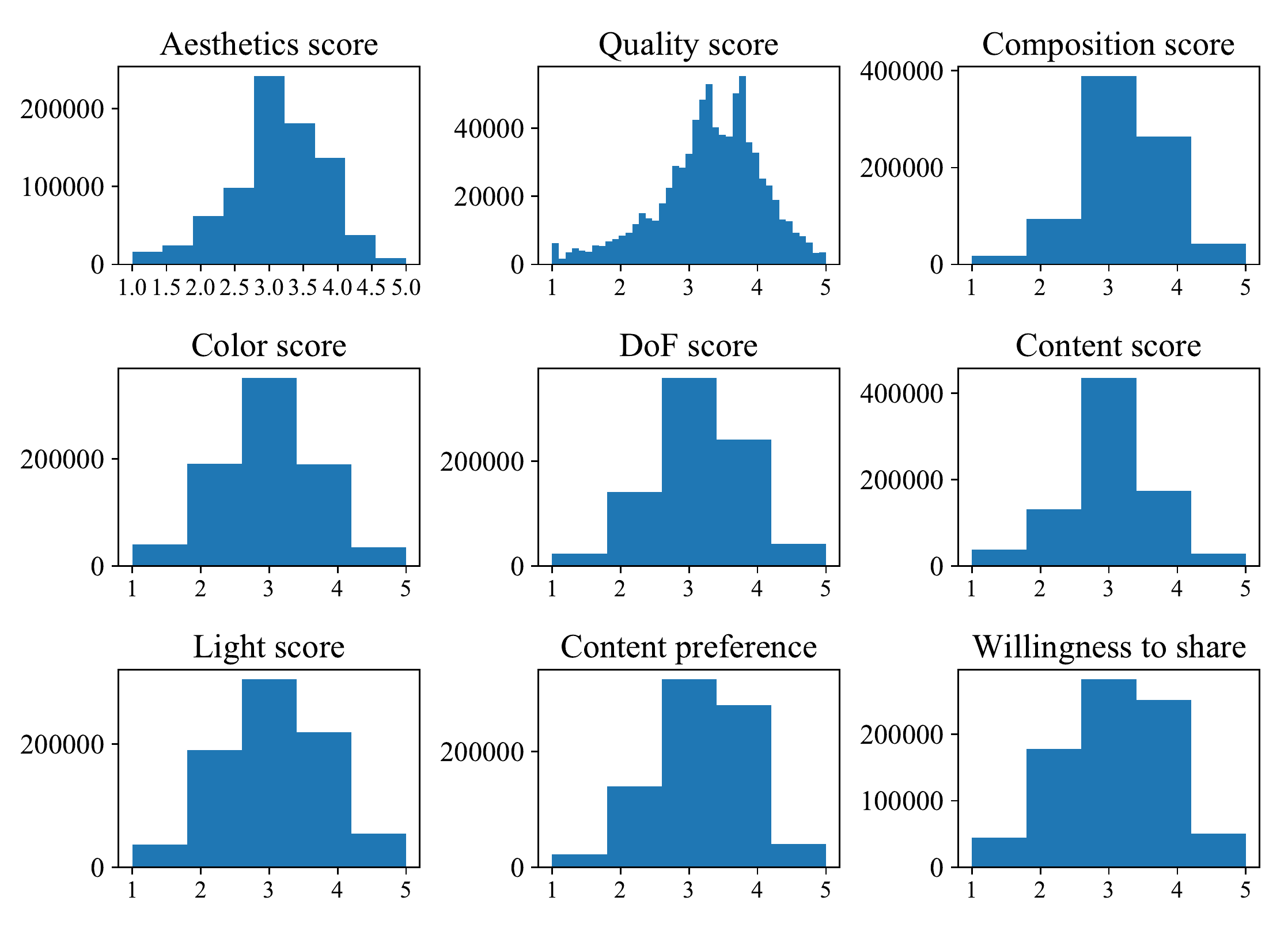}
    \caption{Annotation score distribution. Note that the x axis reflects score of different dimensions, while y axis means the frequency on each score scale.}
    \label{fig4}
\end{figure}

\subsection{Statistical Analysis}
\begin{figure}[htb]
    \centering
    \includegraphics[width=0.4\textwidth]{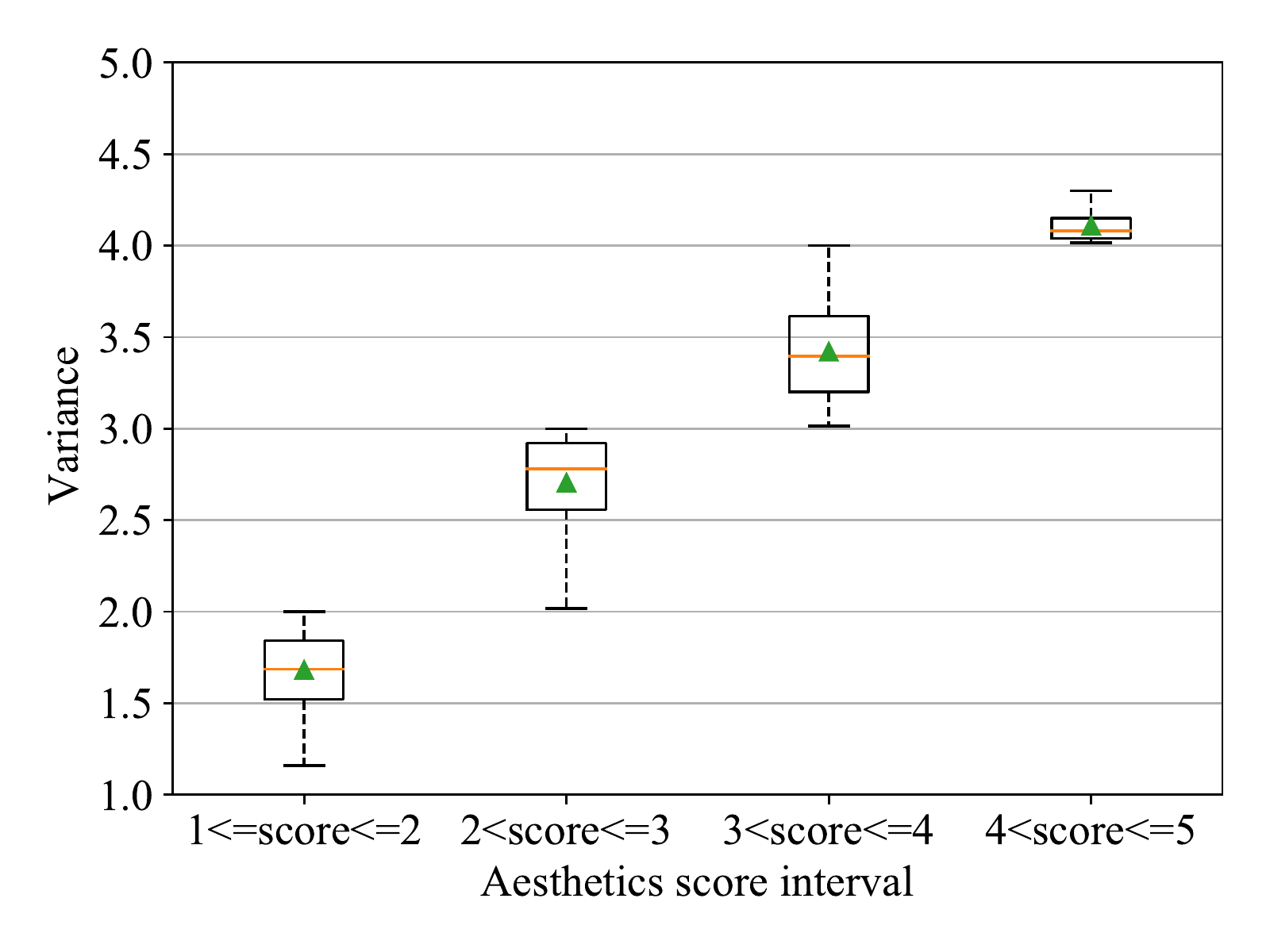}
    \caption{Box plot of different aesthetics score intervals.}
    \label{fig5}
\end{figure}

\paragraph{Annotation distribution} PARA contains rich annotation information. Here, annotation distribution and variance of each dimension are visualized in Figure \ref{fig4} and Figure \ref{fig5}. From Figure \ref{fig4}, it can be observed that the distribution of each attribute is similar. However, they are still slightly different from each other, which indicate that aesthetics attributes are correlated with each other, but still provide unique valuable information. From the box plot of aesthetics score shown in Figure \ref{fig5}, it is easy to observe that in high score interval (4, 5], the aesthetics score does have a lower variance compared with other score intervals. It proves that we do have common cognition on ``what is beautiful". Meanwhile, we do have different aesthetic opinions in other score intervals, such as [1, 2], (2, 3] and (3, 4], which proves the necessity of conducting PIAA research.

\begin{figure}[bht]
    \centering
    \includegraphics[width=0.45\textwidth]{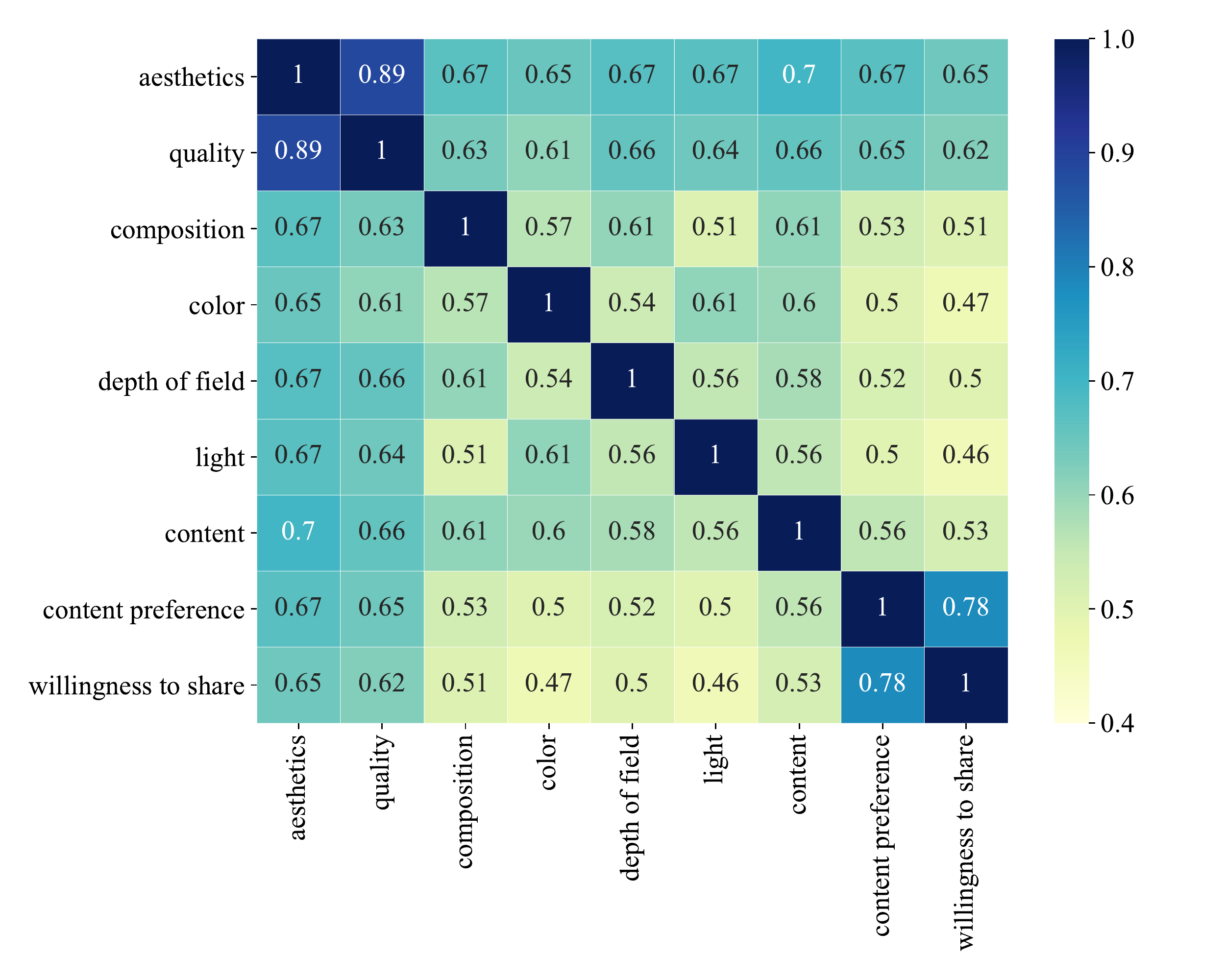}
    \caption{Pearson Linear Correlation Coefficient (PLCC) map among attributes dimensions.}
    \label{fig6}
\end{figure}

\paragraph{Attributes correlation analysis} To understand PARA from the correlation perspective, we visualize the Pearson Linear Correlation Coefficient (PLCC) among each dimension in Figure \ref{fig6}. It can be observed that the correlation between aesthetics and quality is extremely high, which indicates that photo quality can largely affect image aesthetics perception. Meanwhile, correlations among attributes are mostly around 0.5, indicating ``moderately related". It means the annotation information for each dimension contains both commonality and differences. Finally, we notice that the ``content preference" and ``willingness to share" dimensions are also highly correlated, which proves that people are more easily to share a photo when they are enjoying the image content.

\subsection{Subject preference} 
\begin{figure}[bht]
    \centering
    \includegraphics[width=0.47\textwidth]{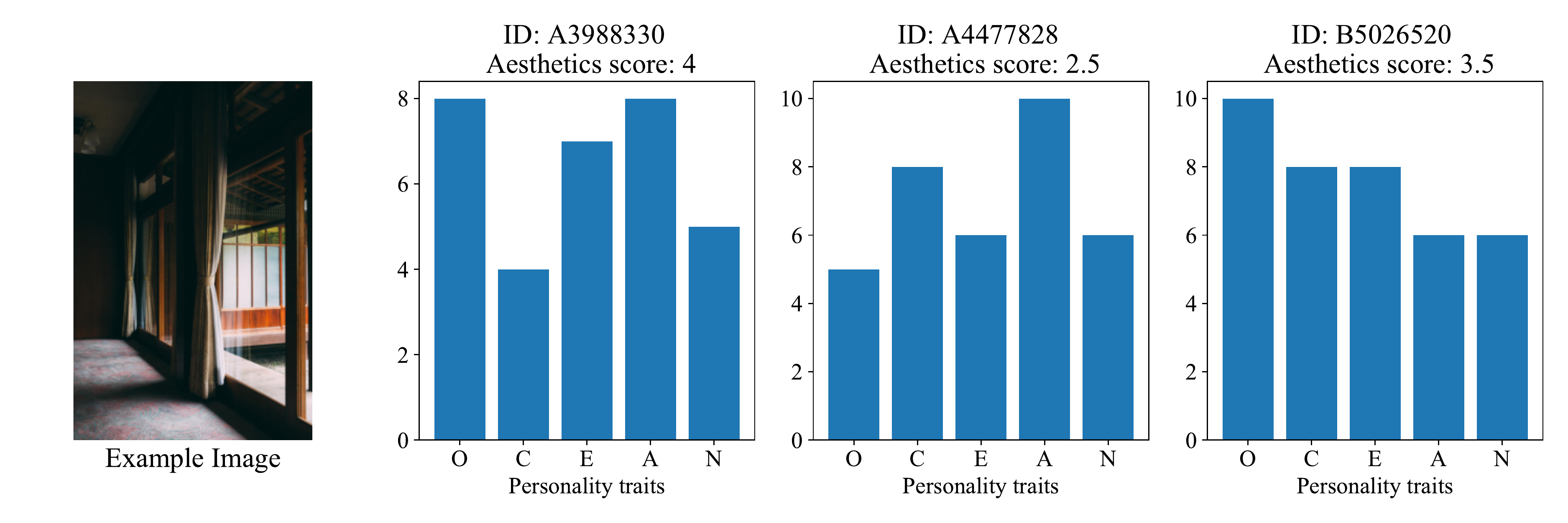}
    \caption{Sample image is rated by three subjects. The related aesthetics scores are 4, 2.5 and 3.5, which is highly differentiated with each other. It is noteworthy that their personality traits are different from each other as well.}
    \label{fig7}
\end{figure}

\begin{figure}[hbt]
    \centering
    \includegraphics[width=0.38\textwidth]{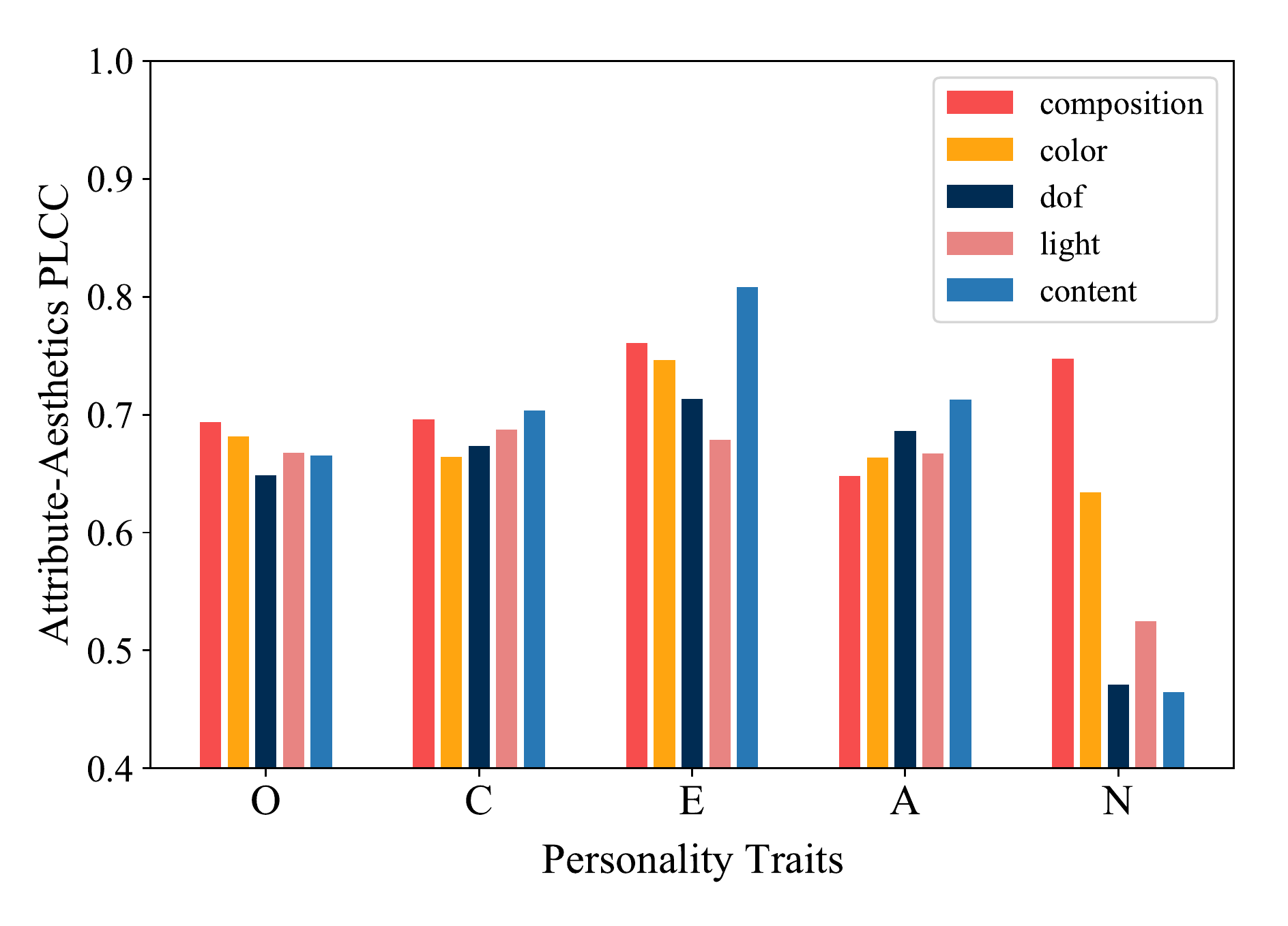}
    \caption{Attributes-aesthetics PLCC with personality traits.}
    \label{fig8}
\end{figure}

Beyond traditional aesthetics judgement dimensions, PARA provides subject portraits as well. Current PIAA algorithms are mostly limited due to the annotation diversity. We believe that subject portrait information can bring more in-depth research opportunities, such as utilizing portrait information by transductive learning \cite{liu2019fewTPN} etc. Therefore, we also collect user portrait information in PARA, such as personality traits and photographic experience. We further study the correlation among personality traits, emotion, annotation difficulty and aesthetics attribute preference to discover characteristics inside. For intuitive observation, we show the personality traits of different subjects together with their aesthetics judgements on sample image in Figure \ref{fig7}. It is easy to observe that subjects with different personality traits have different aesthetic tastes and the given score from three subjects are 4, 2.5 and 3.5.
\paragraph{Personality traits \& attributes preference} We discover correlations between personality, aesthetics score and aesthetics attributes and the result is shown in Figure \ref{fig8}. To obtain the results, we first get the max value among 5 traits and use the associated traits as an accordance to group the data into ``O", ``C", ``E", ``A", ``N". Then, we compute the PLCC between each aesthetics attributes and aesthetics score respectively. Here, it is easy to discover that subjects with high ``N" traits are dissimilar with others. Interestingly, we find that the ``N" traits refer to ``Neuroticism", which means subjects with high ``N" traits tend to over-react to outer stimulation and have stronger emotion reaction compared with others. Similar phenomenon can be found in this work \cite{9599464}. In addition, subjects with high ``E" traits cares more about the content of image when giving the aesthetics judgement, since the PLCC in this dimension has reached over 0.8, indicating ``strongly correlated".

\begin{figure}[htb]
    \centering
    \includegraphics[width=0.38\textwidth]{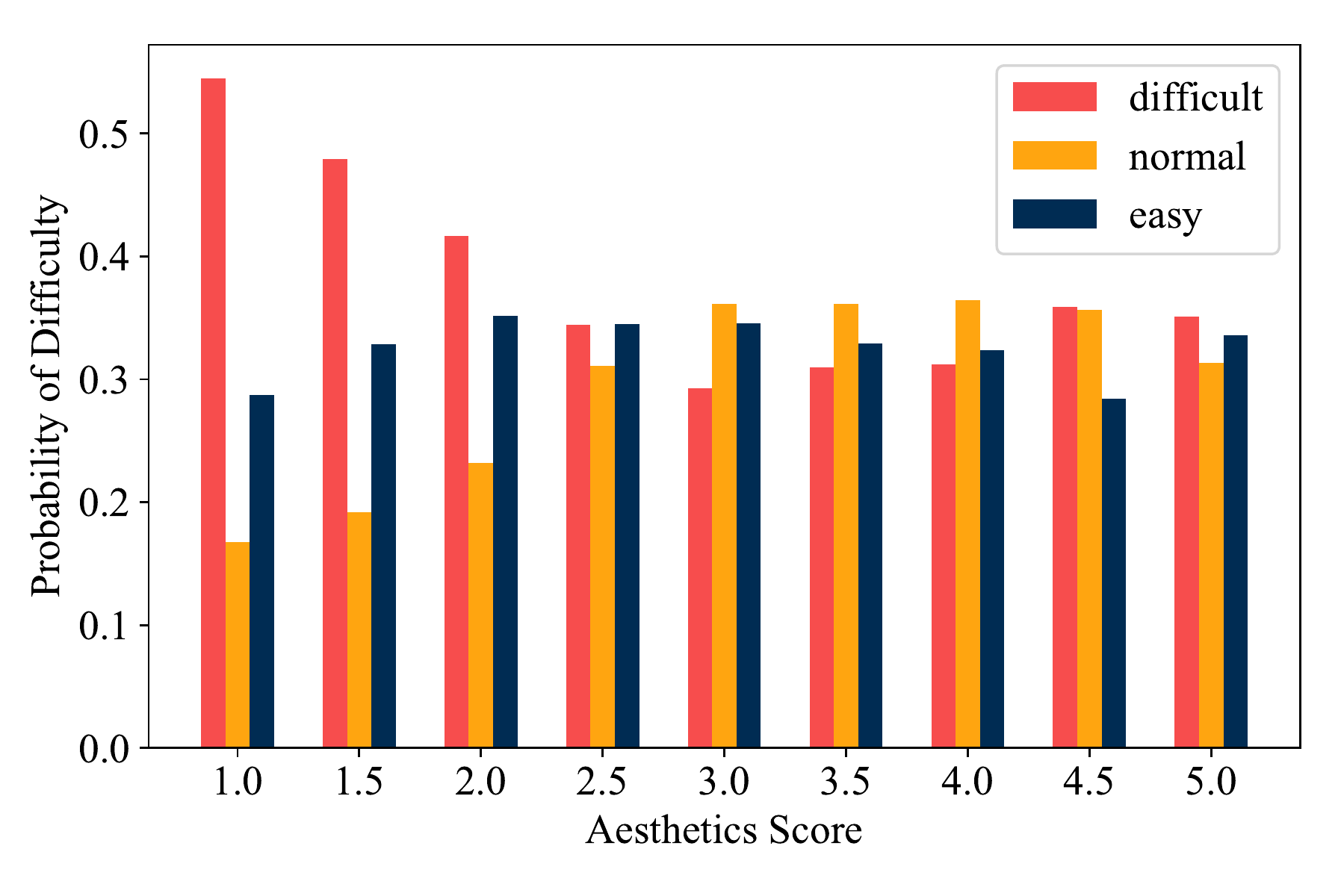}
    \caption{Difficulty of judgement distribution in each aesthetics scores. It can be observed that photo with lower aesthetics score tend to be difficult for subjects to make a judgement.}
    \label{fig9}
\end{figure}

\paragraph{Aesthetics \& difficulty of judgement} We also conduct correlation analysis on ``Difficulty of judgement" and aesthetics score. We calculate the probability of choosing ``difficult", ``normal", and ``easy" at each score and visualize its distribution in Figure \ref{fig9}. It can be observed that subjects feel difficult when evaluating photos with low aesthetics score. To discover reasons behind, we then collect feedback from 10 subjects to discover reasons behind. Interestingly, we notice that subjects say they suffer from dizziness and hard to recognize the scene clearly, so it is hard to give a judgement. This especially happens when annotating aesthetics attributes. In addition, they are more likely to be confused when making a decision between ``worse" and ``much worse", while not when selecting between ``better" and ``much better".


\begin{figure}[htb]
    \centering
    \includegraphics[width=0.38\textwidth]{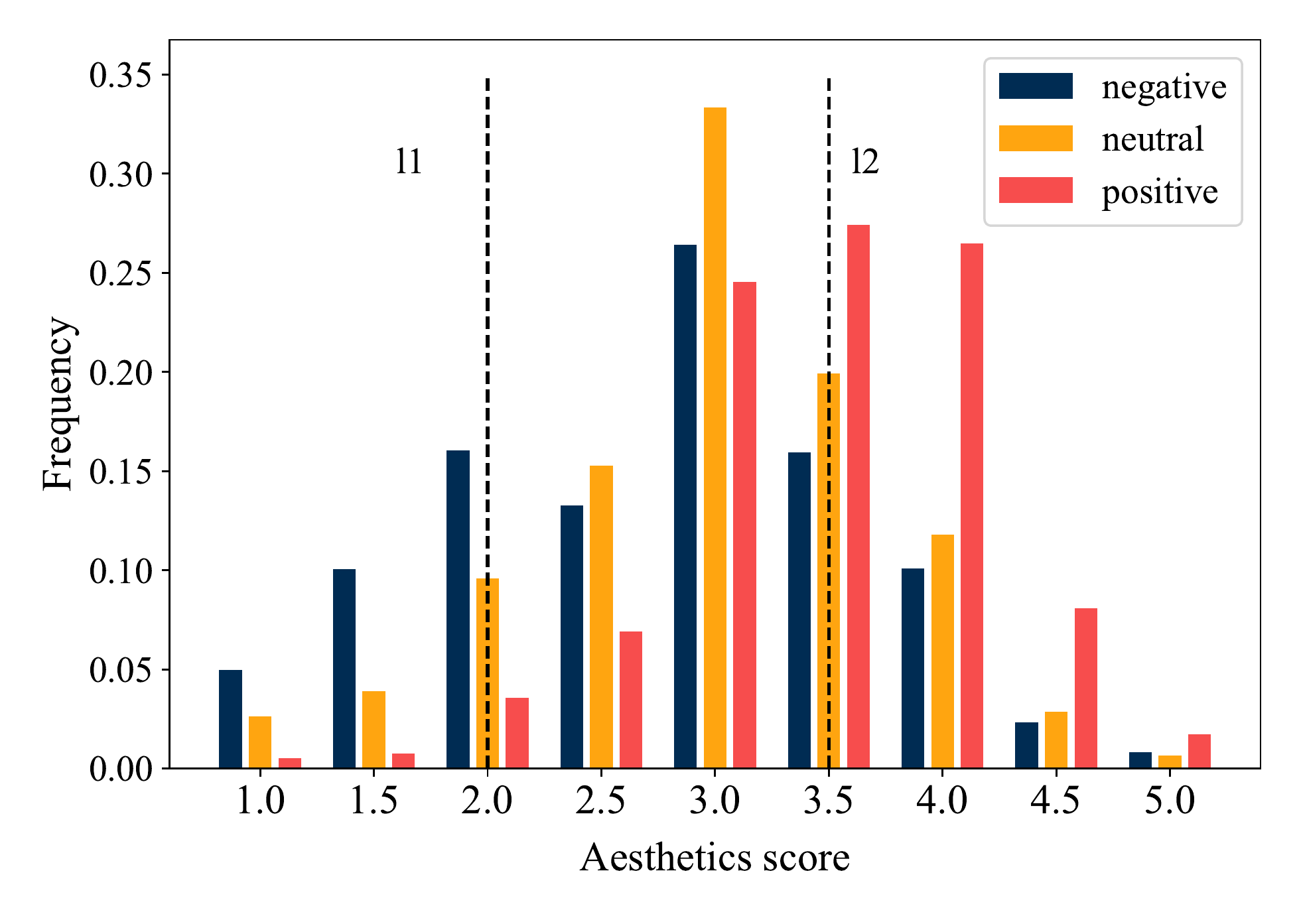}
    \caption{Probabilistic distribution between emotion and aesthetics score.}
    \label{fig10}
\end{figure}

\paragraph{Aesthetics \& emotion}For image emotion dimension, we first group the eight types of emotion into three groups, including ``positive" (amusement, excitement, contentment), ``negative" (disgust, sadness, fear) and ``neutral" (including awe and neutral). Then, we calculate the annotation distribution of grouped emotion categories over aesthetics scores and the related results are shown in Figure \ref{fig10}. Clear conclusion is that images with aesthetics score lower than 2.0 (on the left of l1), are more likely to convey negative emotion. Meanwhile, images with high aesthetics score (over 4.0, on the right of l2), tend to convey positive emotion to subjects. Similar conclusions are also claimed by Cui et al. \cite{yu2019towards}. In their work, they release a conclusion say that if image can arouse positive emotions, they tend to have a higher score; otherwise, they are more likely with low aesthetics score \cite{yu2019towards}.

\begin{figure}[htb]
    \centering
    \includegraphics[width=0.38\textwidth]{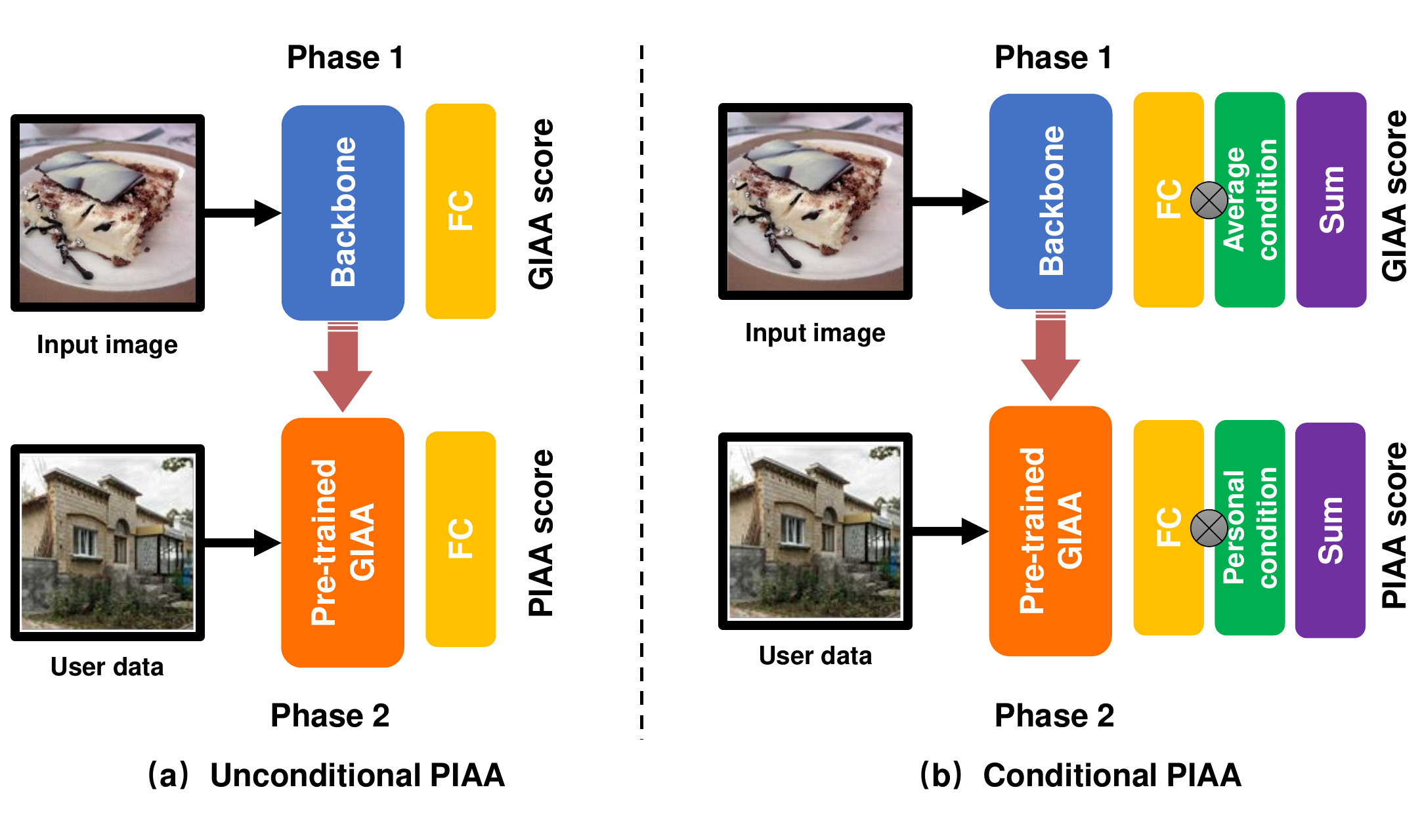}
    \caption{Pipeline of proposed PIAA models. We propose two types of model, including conditional and unconditional PIAA.}
    \label{fig12}
\end{figure}

\section{Benchmark}
\subsection{Conditional and Unconditional PIAA}
To prove the usability and fair comparisons, we conduct a benchmark study on PARA. The proposed benchmark contains two modeling methods, referring to the conditional and unconditional PIAA. The training pipeline of unconditional and conditional PIAA are shown in Figure \ref{fig12} (a) and (b). To implement unconditional PIAA, we first train a GIAA model in phase 1. Then, we directly finetune the GIAA model with personal data to learn personalized preference. As for the conditional PIAA modeling, we add three types of conditional information when modeling, including personality traits, artistic experience and photographic experience in both phase 1 and 2. By multiplying the condition to the last-layer outputs, we learn a ``conditional" GIAA and PIAA model respectively. Finally, we conduct comparison experiment for evaluation. Since PIAA is a typical small sample problem \cite{zhu2020personalized}, we adopt similar experimental settings and evaluation criteria by referring Few-Shot Learning \cite{fe2003bayesian} and previous PIAA research work \cite{Ren_2017_ICCV, zhu2020personalized}. In this work, the experimental settings include 1) without finetune group (``control group"), 2) 10-shot group and 3) 100-shot group. Results are shown in Table \ref{tab:piaa}.

\begin{table*}[htp]
\centering
\resizebox{0.97\textwidth}{!}{
\begin{tabular}{lclcc ccc ccc}
\toprule
\toprule
\multirow{3}*{\textbf{Methods}} & & \multirow{3}*{\textbf{Backbone}} & \multirow{3}*{\textbf{\makecell[c]{Conditional \\information}}} & \multicolumn{3}{c}{\textbf{\emph{SROCC}}}  && \multicolumn{3}{c}{\textbf{\emph{PLCC}}} \\
\cmidrule{5-7}
\cmidrule{9-11}

 & &  & & \textbf{\emph{\makecell[c]{without \\finetune}}} & \textbf{\emph{10 shot}} &\textbf{\emph{100 shot}} & & \textbf{\emph{\makecell[c]{without \\finetune}}} & \textbf{\emph{10 shot}} &\textbf{\emph{100 shot}}\\
\midrule
& & ResNet-18 \cite{he2016deep}   &    /     &$0.6521\pm 0.0038$ &$0.6534\pm 0.0044$&$0.6616\pm 0.0040$& &$0.7069\pm 0.0034$ & $0.7093\pm 0.0044$& $0.7147\pm 0.0031$\\ 
& & MobileNet-V2  \cite{sandler2018mobilenetv2}   &    /     &$0.6696\pm 0.0032$ &$0.6697\pm 0.0031$&$0.6814\pm 0.0041$& &$0.7211\pm 0.0035$ & $0.7214\pm 0.0035$& $0.7302\pm 0.0026$\\   
\makecell[c]{Unconditional \\PIAA group}& & \textbf{ResNet-50*}  \cite{he2016deep}   &    /     &$0.6808\pm 0.0015$ &$0.6811\pm 0.0015$&$0.6952\pm 0.0014$& &$0.7295\pm 0.0014$ & $0.7298\pm 0.0013$& $0.7429\pm 0.0012$\\     
  
& & Swin-TF Tiny  \cite{liu2021Swin}  &    /     &$0.6855\pm 0.0010$ &$0.6859\pm 0.0010$&$0.6988\pm 0.0023$& &$0.7321\pm 0.0012$ & $0.7311\pm 0.0013$& $0.7441\pm 0.0012$\\     
& & Swin-TF Small  \cite{liu2021Swin}    &    /     &$0.6897\pm 0.0013$ &$0.6900\pm 0.0013$&$0.7040\pm 0.0010$& &$0.7354\pm 0.0015$ & $0.7358\pm 0.0015$& $0.7485\pm 0.0011$\\     
\midrule
& & ResNet-50  &    Artistic Exp.     &$0.6854\pm 0.0016$ &$0.6859\pm 0.0016$&$0.6976\pm 0.0012$& &$0.7329\pm 0.0024$ & $0.7332\pm 0.0022$& $0.7419\pm 0.0012$\\   
\makecell[c]{Conditional \\ PIAA group}& & ResNet-50    &    Photographic Exp.  &$0.6826\pm 0.0014$ &$0.6830\pm 0.0014$&$0.6982\pm 0.0010$& &$0.7324\pm 0.0010$ & $0.7326\pm 0.0010$& $0.7447\pm 0.0010$\\    
& & ResNet-50    &    Personality traits     &$0.6908\pm 0.0010$ &$0.6912\pm 0.0009$&$0.7046\pm 0.0015$& &$0.7380\pm 0.0007$ & $0.7384\pm 0.0007$& $0.7509\pm 0.0010$\\     
\bottomrule
\bottomrule
\end{tabular}}
\caption{Experimental results of proposed conditional and unconditional PIAA on PARA. Results of unconditional PIAA with ResNet-50 backbone (marked with * ) are selected as the official benchmark of PARA.}
\label{tab:piaa}
\end{table*}

\paragraph{Implementation details} We randomly select 40 subjects (occupying 10\% of the total subject number) as test subjects. For each subject, 10 and 100 images are randomly selected from his or her personal data as support set and 50 images are also randomly sampled from the rest data as query set. Second, we fine-tune the GIAA model on support set, to refine GIAA into a PIAA model and evaluate performance on query set. Third, to avoid randomness of data selection, data from each subject is sampled and evaluated 10 times and we compute mean value of each evaluation metric. Forth, to observe the robustness and average performance on all test subjects, we repeat the whole pipeline for 10 times and calculate the mean and standard deviation of each evaluation metric over all test subjects as the final results. As for conditional PIAA, we multiply the subjective attributes information to the last-layer output to learn both GIAA and PIAA. Note that we use the ``average subject information" as condition when training GIAA and the rest settings are consistent with the ``unconditional" group.

\subsection{Evaluation Criteria} In this article, we adopt four metrics for GIAA performance evaluation, including Mean Square Error (MSE), Spearman Rank Order Correlation Coefficient (SROCC), Pearson Linear Correlation Coefficient (PLCC) and classification accuracy. While in PIAA, we utilize the SROCC and PLCC as evaluation metrics.

\subsection{Experimental Results}
In this work, the evaluation procedure happens in both GIAA and PIAA. First, experimental results of proposed benchmark is shown in Table \ref{tab:piaa}. We test different GIAA backbones, including ResNet-18 \cite{he2016deep}, ResNet-50 \cite{he2016deep}, MobileNet-V2 \cite{sandler2018mobilenetv2}, Swin-TF tiny \cite{liu2021Swin} and Swin-TF small \cite{liu2021Swin}. We also evaluate the GIAA performance on test set. As for the influence of backbone selection, we also conduct a backbone experiment for comparison. The backbone experimental results are reported in Table \ref{tab:giaabackbone}. 

\begin{table}[!htp]
  \centering
    \resizebox{0.45\textwidth}{!}{
  \begin{tabular}{cccccc}
    \toprule
    \toprule
    \makecell[c]{Conditional \\information}& Backbone & MSE &  SROCC & PLCC & Accuracy\\
    \midrule

       / & ResNet18 \cite{he2016deep}                          & 0.0546     & 0.8538     & 0.9005     & 0.8567 \\
    /    &MobileNet-V2 \cite{sandler2018mobilenetv2}                           & 0.0479     & 0.8706     & 0.9120     & 0.8710 \\
   / & ResNet50 \cite{he2016deep}                                      & 0.0433     & 0.8790     & 0.9208     &  0.8697 \\

    /&Swin-TF Tiny \cite{liu2021Swin}                  & 0.0373    & 0.8971     & 0.9331     & 0.8843 \\
    /&Swin-TF Small \cite{liu2021Swin}                 & 0.0356     & 0.9021     & 0.9355     & 0.8857 \\
       \midrule
    Artistic Exp. & ResNet50                             & 0.0434     & 0.8814     & 0.9206     &  0.8720 \\
     Photographic Exp. &ResNet50                       & 0.0440     & 0.8824     & 0.9215     &  0.8770 \\
     Personality Traits       & ResNet50                          & 0.0416     & 0.8860     & 0.9238     &  0.8763 \\
    \bottomrule
    \bottomrule   
    
  \end{tabular}}
  
  \caption{GIAA backbone experimental results on PARA.}
  \label{tab:giaabackbone}
\end{table}

Experimental results can be summarized as follows. 1) First, we prove that by fine-tuning on 10 and 100 shot personalized data, PIAA can outperform the control group (``without finetune" group). 2) Second, more personalized training data can further improve the finetune performance. In addition, results in 10 shot group are close to the control group, which inspire us to rethink the rationality of 10 shot setting. 3) Third, utilizing subjective attributes information into PIAA modeling can improve model performance compared with the unconditional PIAA group.

\section{Conclusion}
In this article, we have proposed a new PIAA database named ``PARA". The PARA database contains 31,220 images and it is annotated by 438 subjects in total. Rich annotations are attached to each image from 13 dimensions, including 9 image-oriented objective attributes and 4 human-oriented subjective attributes. In addition, desensitized subject information is also provided. Statistical results indicate that the personalized aesthetic preference can be mirrored by the human-oriented subjective attributes. To further prove the value of subject attributes from a computational perspective, we propose a conditional PIAA modeling method by utilizing subject information as conditional prior. Experimental results indicate that adding subjective information can better model personal aesthetic preference, which may bring novel research opportunities for the next-generation PIAA.

{\small

\bibliographystyle{ieee_fullname}
}

\end{document}